\documentclass[journal]{IEEEtran}

\usepackage{cite}
\usepackage{amsmath,amssymb,amsfonts}
\usepackage{algorithmic}
\usepackage{graphicx}
\usepackage{textcomp}
\usepackage{comment}
\usepackage{url}
\usepackage{multirow}

\begin{document}
\title{
Domain Knowledge is Power: Leveraging Physiological Priors for Self‑Supervised Representation Learning in Electrocardiography

}

\author{ 
Nooshin Maghsoodi, Sarah Nassar, Paul F R Wilson, Minh Nguyen Nhat To, Sophia Mannina, Shamel Addas, Stephanie Sibley, David Maslove, Purang Abolmaesumi, and Parvin Mousavi
% \thanks{This paragraph of the first footnote will contain the date on 
% which you submitted your paper for review. It will also contain support 
% information, including sponsor and financial support acknowledgment.  }
\thanks{
This work was supported in part by the Natural Sciences and Engineering Research Council of Canada (NSERC); the New Frontiers in Research Fund (NFRF) through the Social Sciences and Humanities Research Council (SSHRC); and the Vector Institute. Sophia Mannina is supported in part by the Social Sciences and Humanities Research Council. Stephanie Sibley is supported in part by the Canadian Institutes of Health Research (CIHR). David Maslove is supported in part by the Southeastern Ontario Academic Medical Association (SEAMO). The work of Parvin Mousavi was supported in part by a Canada CIFAR AI Chair and a Canada Research Chair.
\textit{(Corresponding Author: Parvin Mousavi, e-mail: mousavi@queensu.ca)}}
% \thanks{This study was approved by Queen’s University Health Sciences Research Ethics Board—Reference number 6024689. Consent in this study was waived because the data were collected as part of routine care and were stored in a de-identified format.}
% \thanks{Stephanie Sibley declares the following conflicts of interest: receipt of honoraria from Think Research; meeting sponsorships from Boston Scientific, Trimedic, and Icentia; and service as a hospital organ-donation physician with the Trillium Gift of Life Network at Ontario Health.}
\thanks{Nooshin Maghsoodi and Paul F R Wilson are with the School of Computing, Queen’s University, Kingston, ON, Canada. }% <-this % stops a space
\thanks{Sarah Nassar is with the Department of Electrical and Computer Engineering at Queen's University, Kingston, ON, Canada.}% <-this % stops a space
\thanks{Minh Nguyen Nhat To is with the Department of Electrical and Computer Engineering at the University of British Columbia, Vancouver, BC, Canada, and Vector Institute, Toronto, Canada.}
\thanks{Sophia Mannina and Shamel Addas are with the Smith School of Business at Queen's University, Kingston, ON, Canada.}% <-this % stops a space
\thanks{David Maslove is with the Departments of Medicine and the Department of Critical Care Medicine at Queen's University, Kingston, ON, Canada.}% <-this % stops a space
\thanks{Stephanie Sibley is with the Department of Emergency Medicine and the Department of Critical Care Medicine at Queen's University, Kingston, ON, Canada.}
\thanks{Purang Abolmaesumi is with the Department of Electrical and Computer Engineering at the University of British Columbia, Vancouver, BC, Canada.}% <-this % stops a space
\thanks{Parvin Mousavi is with the School of Computing, Queen’s University, Kingston, ON, Canada and Vector Institute, Toronto, Canada.}% <-this % stops a space
}

\maketitle

\begin{abstract}

\textit{Objective:} Electrocardiograms (ECGs) play a crucial role in diagnosing heart conditions; however, the effectiveness of artificial intelligence (AI)-based ECG analysis is often hindered by the limited availability of labeled data. Self-supervised learning (SSL) can address this by leveraging large-scale unlabeled data. We introduce PhysioCLR (Physiology-aware Contrastive Learning Representation for ECG), a physiology-aware contrastive learning framework that incorporates domain-specific priors to enhance the generalizability and clinical relevance of ECG-based arrhythmia classification.~\textit{Methods:} During pretraining, PhysioCLR learns to bring together embeddings of samples that share similar clinically relevant features while pushing apart those that are dissimilar. Unlike existing methods, our method integrates ECG physiological similarity cues into contrastive learning, promoting the learning of clinically meaningful representations. Additionally, we introduce ECG-specific augmentations that preserve the ECG category post-augmentation and propose a hybrid loss function to further refine the quality of learned representations.~\textit{Results:} We evaluate PhysioCLR on two public ECG datasets, Chapman and Georgia, for multilabel ECG diagnoses, as well as a private ICU dataset labeled for binary classification. Across the Chapman, Georgia, and private cohorts, PhysioCLR boosts the mean AUROC by 12\% relative to the strongest baseline, underscoring its robust cross-dataset generalization.~\textit{Conclusion:} By embedding physiological knowledge into contrastive learning, PhysioCLR enables the model to learn clinically meaningful and transferable ECG features. ~\textit{Significance:} PhysioCLR demonstrates the potential of physiology-informed SSL to offer a promising path toward more effective and label-efficient ECG diagnostics.
\end{abstract}


% \begin{Abbreviations and Acronyms}
% e
% \end{Abbreviations and Acronyms}

\section{Introduction}
\label{sec:introduction}

Deep learning (DL) has driven substantial progress in biomedical signal analysis~\cite{faust2018review}. %Compared to traditional feature engineering approaches, 
Modern DL networks are capable of learning discriminative features directly from high-dimensional raw data and scaling to very large datasets. Successive architectural innovations, coupled with increases in depth and number of parameters, have yielded models of ever-increasing capacity and expressiveness. However, these gains in representational power incur a proportional increase in the amount of data required to train these models effectively. Because acquiring labels for biomedical signals is typically time-consuming and expert-dependent, the problem of \emph{label scarcity} presents a major barrier to achieving further performance improvements. 

This challenge is exemplified in electrocardiogram (ECG) analysis. In this setting, DL models provide a promising tool to enable automatic and high-throughput monitoring and diagnosis of heart conditions, including arrhythmias, myocardial infarction, and conduction disorders. %This includes the potential to identify subtle signals in ECG which elude human interpretation~\cite{}
ECG is cost-effective and non-invasive, making it relatively straightforward to obtain large amounts of data; however, pathological events occupy only a tiny fraction of typical recordings and demand meticulous expert annotation, creating a severely label-scarce training regime. In this work, our aim is to improve the automatic diagnosis of heart conditions using ECG recordings by learning better representations through self-supervised learning. By doing so, we seek to enhance arrhythmia classification and detection of abnormal rhythms in both standard clinical ECGs and challenging ICU settings, where signals are often noisier.

Self-supervised learning (SSL) offers a potential solution for label scarcity. SSL algorithms involve designing a pretraining task that does not depend on labels, but which forces the network to map low-level signal patterns to high-level semantics relevant for downstream applications. 
Using this task, a model can be pretrained on large-scale unlabeled datasets, and then can be finetuned for a downstream task using a much smaller number of labeled samples. Tasks such as \emph{alignment}, where models are trained to produce similar representations for semantically similar pairs of inputs, (termed \emph{positive pairs})~\cite{chen2020,byol,dinov2}; and \emph{reconstruction}, where models are trained to reconstruct an original input given a partially masked or corrupted version of it~\cite{reconst1, reconst2, reconst3}, have proven to be capable of driving powerful representation learning~\cite{dinov2,mohamed2022review}. %In recent years, SSL has demonstrated performance comparable to, and in some cases exceeding, that of fully supervised methods in both computer vision~\cite{dinov2} and speech~\cite{mohamed2022review}. %proving its capacity to extract rich structure from high‑dimensional continuous signals. 
This success strongly motivates its adoption for biomedical signals such as ECG.

Despite its promise, applying SSL to biosignals such as ECG is challenging because SSL hinges on modality-specific design choices. For example, alignment objectives require positive pairs created by appropriate augmentations or sampling (e.g., crops/flips for images, orthogonal views in chest X-ray), while reconstruction objectives need masking schemes tailored to the signal structure (e.g., word-level masking in text). These components are typically designed based on domain knowledge and extensive experimentation, with inappropriate or suboptimal choices significantly degrading the quality of learned representations~\cite{augmentationIssue, Gopal2021}. The success of SSL in ECG analysis improves when methods are guided by domain knowledge and remain faithful to the underlying physiology of the signals.

Recognizing this need, recent studies have sought to add physiological priors into SSL for ECG. ECG-specific data augmentations have been proposed~\cite{mehari2022,Gopal2021, Le2023,chen2021}, together with sampling strategies such as using different segments from the same patient~\cite{Kiyasseh2021,Wang2024,Oh2022}, to improve the generation of positive pairs for alignment. %have been proposed with the assumption that these segments may share the same pathologies. 
ECG-specific masking strategies have been designed to improve the generation of effective reconstruction targets~\cite{chen2024multi,na2024guiding}. %Strategies such as lead-aware~\cite{chen2024multi} or beat-aligned~\cite{na2024guiding} masking have been designed to improve on generic reconstruction objectives. 
Some works add a small number of physiologically derived features either as auxiliary prediction targets~\cite{Liu2024} or as extra encoder inputs~\cite{Zhou2025}. 

Despite these promising developments, prior work suffers from two key limitations. First, existing methods are typically fragmented, addressing isolated components such as augmentations, sampling, or reconstruction in isolation, rather than through a unified framework. Second, most approaches leverage only narrow aspects of ECG physiology, leaving a broad spectrum of clinically relevant features underutilized. A more comprehensive and physiologically grounded design is needed to fully realize the potential of SSL for ECG.

%Despite these improvements, these methods typically only focus on a relatively narrow scope of ECG physiology within their designs. This leaves a wide breadth of domain knowledge, including physiologically and clinically relevant signal properties such as waveform, rhythm, and hemodynamic features, untapped.

%at most one or two aspects of ECG physiology, 

%Redesigned augmentations to preserve semantics of ECG signals (e.g., semantic peaks noise smoothing and split-join transformation~\cite{16}), and strategies such as sampling segments from the same patient and based on temporal proximity~\cite{20, 21, 22}, have been introduced to improve the generation of positive pairs for alignment. Strategies such as lead-aware~\cite{chen2024multi} or beat-aligned~\cite{na2024guiding} masking improve on generic reconstruction objectives. Some works add a small number of physiologically derived features, such as RR intervals and P-wave statistics, either as auxiliary prediction targets~\cite{24} or as extra encoder inputs~\cite{23}. 

%Despite their success, these methods remain fragmented, typically only addressing a single facet of SSL methodology and using at most a small number of the wide spectrum of waveform, rhythm, and hemodynamic features available as physiological priors.

In this study, we propose \textbf{PhysioCLR}, a comprehensive and unified method to exploit physiological priors in SSL for ECG.  Our specific contributions tailored to the core clinical tasks of ECG interpretation are as follows: 

\begin{enumerate}

\item %We design a novel, unified SSL framework for ECG, which--to the best of our knowledge--makes the most extensive use of domain knowledge to date, leveraging more than 100 physiological features across several categories to learn robust, clinically meaningful representations.

%We design a novel, unified SSL framework for ECG that, to the best of our knowledge, makes the most comprehensive use of domain knowledge to date—leveraging a diverse set of physiological features that capture key morphological and temporal differences across arrhythmias to learn robust, clinically meaningful representations.

%We present the first comprehensive SSL framework for ECG that integrates physiological priors across all key design dimensions. Unlike prior works that focus on isolated components (e.g., augmentations or sampling), our method unifies alignment and reconstruction objectives under a single framework, guided by over 100 physiological features spanning morphological, temporal, and rhythm-based characteristics.

We propose the first self-supervised learning framework for ECG that systematically integrates physiological priors across all key design components—sample selection, data augmentation, and reconstruction. In contrast to prior work, which typically addresses these aspects in isolation, our method unifies alignment and reconstruction objectives within a single, principled framework. The design is informed by over 100 diverse physiological features encompassing morphological, temporal, rhythmic, and hemodynamic characteristics, enabling the learning of robust and clinically meaningful representations.

\item We introduce three physiologically informed components to enhance representation learning:
(i) a sample selection strategy based on biological similarity derived from a comprehensive set of physiological signal features,
(ii) a peak-aware reconstruction loss that emphasizes diagnostically important waveform regions, and
(iii) a heartbeat-shuffling augmentation to promote temporal robustness.
These components are integrated into a hybrid self-supervised objective that combines a contrastive loss with an auxiliary reconstruction term, enabling the model to capture both semantic similarity and fine-grained waveform structure.
% We proposed a heartbeat-shuffling augmentation that reorders complete heartbeats within ECG segments to introduce variability, thereby encouraging the model to learn representations that are robust to changes in heartbeat sequence order while preserving the original class label.

%We propose key methodological components--a sample selection strategy based on semantic similarity of ECG samples according to their biological characteristics, a peak-aware reconstruction loss, and a heartbeat-shuffling augmentation strategy--and demonstrate how these components can be integrated into a unified SSL objective that co-optimizes alignment and reconstruction terms. 

\item We demonstrate that our physiology-informed SSL pretraining method learns more transferable and clinically relevant representations than prior approaches. Across the public PhysioNet 2021 dataset and the private KGH ICU dataset, these representations lead to improved performance on downstream tasks, including multilabel arrhythmia classification and binary atrial fibrillation (AFib) detection.

\end{enumerate}

\section{Related Work}
\subsection{ECG Physiology and Signal Characteristics}
\label{subsection:ecgIntro}
The interpretation of ECG signals depends on analyzing both morphological and temporal features that reflect the underlying electrophysiological activity of the heart. A regular heartbeat consists of a series of distinguishable peaks, namely the P-wave, QRS complex, and T-wave, each corresponding to specific phases of a single heartbeat. These peaks, along with the intervals between them, provide the foundation for clinical assessment. For instance, the number and amplitude of each peak type indicate the presence and strength of atrial and ventricular activity, while the time intervals between peaks (such as RR and QT intervals) help assess rhythm regularity. In particular, heart rate variability (HRV), calculated from RR intervals, is a key indicator in detecting and differentiating arrhythmias. Other features, including wave durations, slope characteristics, and overall signal energy, capture the dynamics and intensity of electrical activity across the heartbeat. Together, these morphological and temporal descriptors support both manual and automated ECG interpretation by highlighting clinically relevant patterns linked to a broad spectrum of cardiac conditions~\cite{clifford2006advanced,acharya2017automated, Surawicz}.

\subsection{Machine Learning for ECG Analysis}

Machine learning has been investigated for a wide range of ECG analysis tasks. These include arrhythmia classification~\cite{hannun2019cardiologist}, rhythm abnormality detection such as atrial fibrillation~\cite{clifford2017afdb}, myocardial infarction diagnosis~\cite{acharya2017deep}, and beat segmentation~\cite{ibtehaz2022ecg}. Other applications include disease progression monitoring, patient risk stratification, and biometric identification~\cite{acharya2017deep, ibtehaz2022ecg}.

Early ECG analysis methods relied on manually engineered features combined with traditional machine learning algorithms such as support vector machines and $K$-nearest neighbors for analysis~\cite{svm, svm2, svm3}. Although effective on curated datasets, these approaches lacked scalability and struggled to generalize to diverse patient populations. Deep learning has since become the dominant paradigm, enabling the learning of rich features directly from raw ECG waveforms. 

The design of effective network architectures to learn these feature embeddings has been the subject of many studies. Convolutional neural networks (CNNs) have been widely adopted due to their ability to extract local waveform features from short signal segments~\cite{cnn1, cnn2, cnn3}. However, the limited receptive field of CNNs makes them less effective at capturing longer-range temporal dependencies, which are critical for detecting rhythm abnormalities. Conversely, attention-based networks such as transformers~\cite{transformer, transformer2, ecgfm} excel at capturing long-range dependencies. Most current state-of-the-art networks adopt a hybrid network architecture consisting of an initial CNN stage to learn a local signal representation, followed by a transformer stage to aggregate a global representation~\cite{Oh2022, ecgfm}. %In this work we follow this successful architectural design.

\subsection{Self-Supervised Learning for ECG Analysis}

Due to its promising ability to learn strong feature embeddings of data directly without the requirement of labels, the development of SSL for ECG analysis is an active area of research. 

\textit{Alignment-Based Methods:} Aligning features of semantically similar pairs of data is proven to be a powerful concept in SSL, and underlies the success of methods including contrastive learning~\cite{chen2020}, non-contrastive learning~\cite{byol}, and self-distillation~\cite{dinosr}. In particular, \emph{contrastive learning}, which aims to align semantically similar data (positive pairs) while pushing apart semantically different data (negative pairs), has gained significant attention in ECG analysis. 

\begin{figure*}[hbtp]
\centerline{\includegraphics[width=\textwidth]{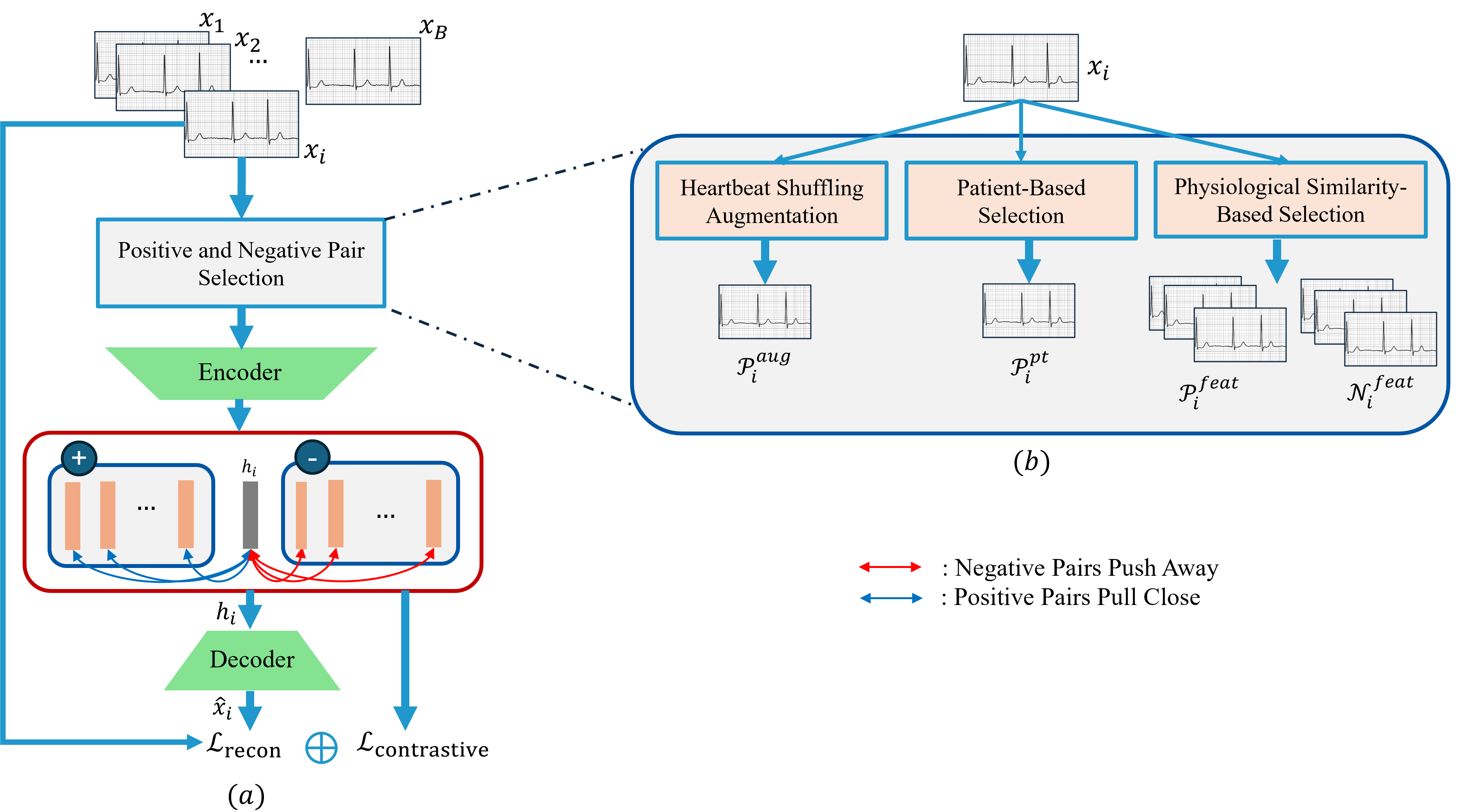}}
    \caption{
Overview of the proposed model: (a) Each ECG segment, $x_i$, is selected as the anchor segment, and a set of positive pairs and a set of negative pairs for this segment are selected. Then, all the positive and negative samples, along with $x_i$ itself, are encoded. In the embedding space, the contrastive loss aims to bring $x_i$ closer to its positive pairs while pushing it farther from its negative pairs. Additionally, the decoder reconstructs $x_i$ from $h_i$ and compares the peaks between the original signal and the reconstructed one. 
(b) Inside the positive and negative pair selection component, for each $x_i$, heartbeat shuffling augmentation generates a positive sample. Patient-level positive pair selection chooses another positive sample based on time adjacency. Feature-level pair selection selects additional positive pairs and also identifies negative pairs based on ECG features.
Compared to common contrastive learning methods, our approach introduces modifications in both positive and negative pair selection to learn class-specific and physiologically meaningful representations. Additionally, we introduce a hybrid loss function that combines contrastive learning objectives with reconstruction-based objectives to improve representation quality further.
% Our approach enhances positive and negative pair selection for class-specific, physiologically meaningful representations and integrates a hybrid loss function by adding a decoder to the model.
    }
    \label{fig:overview}
\end{figure*}

The effectiveness of contrastive learning depends on how positive and negative pairs are selected. Techniques based on sampling and data augmentation are both common in SSL literature and have been adopted for ECG. For augmentations, several studies~\cite{mehari2022,Gopal2021, Le2023,chen2021,Wang2023} have introduced ECG-aware augmentations. Importantly, such augmentations should ensure that physiological and temporal features of ECG segments are maintained, so that class labels remain unchanged. In our work, we propose an augmentation method that explicitly respects these properties, promoting more generalizable contrastive representations.

% that aim to preserve the waveform morphology and temporal consistency of cardiac signals. 

For sampling, a common approach in the ECG domain is \emph{patient-based} pair selection, where temporally adjacent ECG segments from the same patient are treated as positive pairs, and segments from different patients are treated as negative pairs~\cite{Kiyasseh2021}. This approach has been widely adopted in subsequent ECG contrastive learning studies~\cite{Wang2024,Oh2022,ecgfm}.

While successful, these strategies for pair selection induce a strong risk of \emph{false-negative} pairs~\cite{huynh2022boosting}: For example, pairs of data representing the same pathology could be incorrectly assigned as negative pairs because they come from different patients. Moreover, positive pairs based on temporal adjacency within the same patient may not include diverse examples of similar cardiac conditions across different patients, limiting the model’s ability to learn generalizable features.  Resolving this risk through the introduction of a sampling strategy more faithful to the physiological similarity of ECG segments is an important contribution of our work.  

\textit{Reconstruction-Based Methods:} Reconstruction approaches train a network to predict the original values of masked or corrupted input segments, and include variants such as autoencoders~\cite{reconst1}, predictive coding~\cite{reconst2}, and masked-signal modeling~\cite{reconst3}. In ECG, approaches such as masked autoencoders~\cite{masked1,masked2} have been adapted to reconstruct waveform segments from context, capturing rhythm and global morphology. However, standard masking strategies may cause the model to focus on unimportant low-level signal reconstruction while ignoring clinically important fine-grained features (e.g., subtle changes in the P-wave or QRS complex). Addressing this limitation through physiologically informed reconstruction tasks is a key contribution of our work.

Some recent works use ECG-specific knowledge to improve learning. For example, Zhu et al.~\cite{Zhou2025} added tasks to model RR irregularity and missing P-waves for atrial fibrillation detection, while still using positive pairs from the same ECG and negatives from different ECGs in contrastive learning. Liu et al.~\cite{Liu2024} proposed Morphology-Rhythm Contrastive Learning (MRC), which represents heartbeat shape using a single beat and rhythm using a binary pulse signal that marks the R-peak positions. Each ECG is paired with its beat and pulse as positive examples, while beats and pulses from other ECGs are used as negatives.

%To leverage the complementary strengths of contrastive and reconstruction-based learning, hybrid self-supervised methods jointly optimize both objectives. For example, Xu et al.~\cite{rebar} adds a reconstruction term to contrastive learning for general time-series data, while Phim et al.~\cite{cmelt} blend masked autoencoding with contrastive learning in an ECG-language pretraining framework. Although such methods improve robustness and generalization, preserving physiologically significant signal properties, such as peak locations, waveform intervals, and timing patterns, that are essential for clinical interpretation and diagnostic accuracy, is missed.

 These advances highlight the rich interplay between physiological insights and deep learning for ECG interpretation. Building on this foundation, we introduce PhysioCLR: a unified self-supervised learning framework that combines feature-informed sampling, physiology-aware augmentations, and peak-level reconstruction. By integrating these components, PhysioCLR learns clinically meaningful representations, enhancing arrhythmia classification from ECG recordings.

\section{Methodology}
\label{sec:method}
 Our method is designed to learn clinically relevant ECG representations in a self-supervised manner by integrating three key components: feature-informed positive and negative pair selection, ECG-specific data augmentation, and a reconstruction objective that emphasizes physiologically important waveform regions. Fig.~\ref{fig:overview} presents an overview of this approach. The overall goal is to enable accurate classification of arrhythmias across diverse ECG datasets, from public 12-lead recordings to 4-lead ICU data. In part (a), the overall training pipeline is illustrated: each ECG segment is passed through a shared encoder alongside selected positive and negative samples. The generated embeddings are then used to compute the contrastive loss, while a decoder reconstructs the original signal to compute the reconstruction loss. Part (b) highlights the three mechanisms used for generating positive and negative pairs: patient-based temporal adjacency, physiology-informed similarity, and heartbeat-shuffling augmentation. Together with the decoder, these components contribute to the overall self-supervised training objective described in the subsequent sections.

\subsection{Contrastive Loss}
\label{subsec:contrastive}
\subsection{Problem Formulation}
Let $\mathcal{X} = \{\mathbf{x}_i\}_{i=1}^N$ denote a dataset of unlabeled ECG segments, where each $\mathbf{x}_i \in \mathbb{R}^{C \times T}$ represents a multichannel time series with $C$ leads and $T$ time points. The goal is to learn an encoder $f_\theta: \mathbb{R}^{C \times T} \rightarrow \mathbb{R}^d$ that maps input segments to latent representations $\mathbf{h}_i = f_\theta(\mathbf{x}_i)$. 

The encoder parameters $\theta$ are optimized by minimizing a self-supervised objective:
\[
\theta^\ast = \arg\min_\theta \mathcal{L}_{\text{SSL}}(\theta).
\]

The loss $\mathcal{L}_{\text{SSL}}$ comprises two components:
\[
\mathcal{L}_{\text{SSL}} = \mathcal{L}_{\text{contrastive}} + \lambda \mathcal{L}_{\text{recon}},
\]
where $\mathcal{L}_{\text{contrastive}}$ encourages representations of semantically similar inputs to be close in the embedding space, $\mathcal{L}_{\text{recon}}$ enforces signal-level fidelity through reconstruction, and $\lambda$ is a hyperparameter controlling the relative weight of the terms. We now describe each component in detail.

For a given training example (called the \emph{anchor}), the contrastive loss is computed by comparing its embedding to embeddings of other samples. Specifically, a set of \emph{positive} (semantically similar) and \emph{negative} (semantically different) pairs is constructed, and the loss encourages the embedding to be similar to those of positive pairs while being different from negative pairs. Formally, let ${\{\mathbf{x}_i}\}_{i=1}^{B}$ denote a batch of anchor samples: we define three distinct mechanisms for selecting positives and negatives, then describe the full loss computation based on these sets.
\begin{figure*}[!t]
    \centering
    \includegraphics[width=\textwidth]{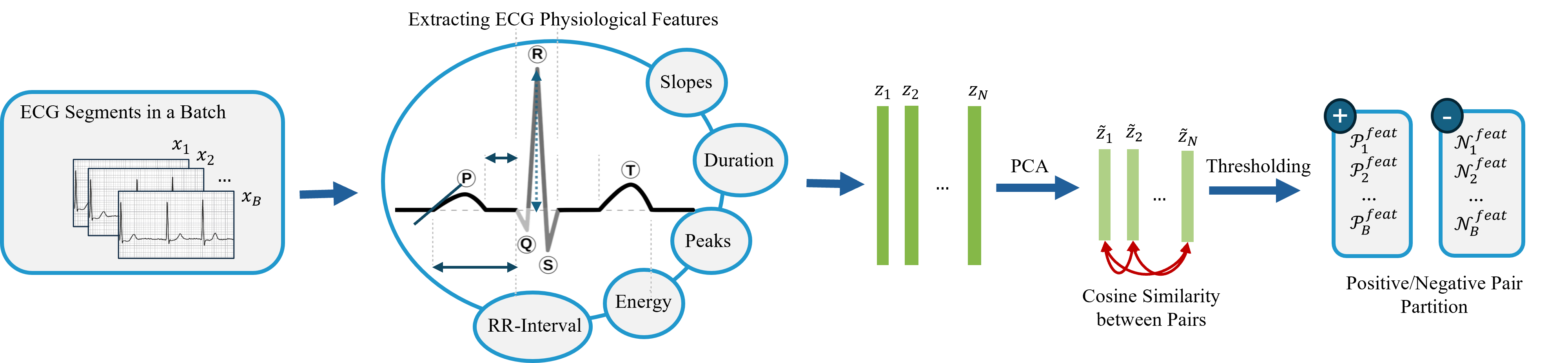}
    \caption{
Feature-Level Sample Selection: This diagram shows how contrastive pairs are created using ECG-specific features for self-supervised learning. Each ECG segment is first analyzed to extract detailed morphological and temporal features—such as slopes, durations, peak counts, energy, and RR-intervals. These features are combined into vectors and reduced in size using PCA. Cosine similarity comparisons of these vectors then identify similar (positive) and dissimilar (negative) pairs for contrastive training.
    }
    \label{fig:class_aware_ss}
\end{figure*}

\subsubsection{Patient-Based Positive Pair Selection}

Here, we generate positive pairs by using patient identity and temporal continuity as a proxy for semantic similarity. It follows the Contrastive Multi-segment Coding (CMSC) framework~\cite{Kiyasseh2021}. Let $\tilde{\mathbf{x}} \in \mathbb{R}^{C \times 2T}$ be a 10-second ECG segment. We partition it into two contiguous 5-second subsegments:
\[
\tilde{\mathbf{x}} = [\mathbf{x}_a, \mathbf{x}_p], \quad \mathbf{x}_a, \mathbf{x}_p \in \mathbb{R}^{C \times T}.
\]
Here, $\mathbf{x}_a$ serves as the \textit{anchor} and $\mathbf{x}_p$ as the \textit{positive} sample. These segments form a positive pair $(\mathbf{x}_a, \mathbf{x}_p)$ based on local temporal continuity. Over a batch $\{ \tilde{\mathbf{x}}_i \}_{i=1}^B$, this yields $\mathcal{P}^{\text{pt}}_i = \{ \mathbf{x}_p^{(i)} \}$ for each anchor $\mathbf{x}_a^{(i)}$.

\subsubsection{Physiological Similarity-Based Selection}
\label{subsection:feature-level}

This component, which is illustrated in Fig.~\ref{fig:class_aware_ss}, generates positive pairs based on the physiological similarity of segments. We extract hand-crafted physiologically meaningful features $\mathbf{z}_i \in \mathbb{R}^F$ from each ECG segment. For each segment, up to 
150 features are extracted using peak detection and morphology metrics (e.g., peak counts, amplitudes, durations) provided by the Pan-Tompkins algorithm and \texttt{NeuroKit2}~\cite{neurokit2}. %\footnote{Full details provided in Appendix XXX and in our source code}.%
 These features are then zero-padded to a fixed length, normalized, and projected to a lower-dimensional space using PCA:
\[
\tilde{\mathbf{z}}_i = \text{PCA}(\text{Norm}(\text{Pad}(\mathbf{z}_i))).
\]
The PCA transformation is precomputed on the entire training set and subsequently applied to each sample during training.

We use the cosine similarity $\text{sim}(\cdot, \cdot)$ of the features $\tilde{\mathbf{z}}_i$ as a mechanism to compare the physiological similarity of samples. Based on this similarity, we define the feature-based positive and negative sets as:
\[
\mathcal{P}_i^{\text{feat}} = \left\{ \mathbf{x}_j \mid \text{sim}(\tilde{\mathbf{z}}_i, \tilde{\mathbf{z}}_j) \geq \delta \right\}, \quad
\mathcal{N}_i^{\text{feat}} = \left\{ \mathbf{x}_k \mid \text{sim}(\tilde{\mathbf{z}}_i, \tilde{\mathbf{z}}_k) < \delta \right\},
\]
where $\text{sim}(\tilde{\mathbf{z}}_i, \tilde{\mathbf{z}}_j) = \frac{\tilde{\mathbf{z}}_i^T\tilde{\mathbf{z}}_j}{||\tilde{\mathbf{z}}_i|| ||~\tilde{\mathbf{z}}_j||}.$ and $\delta$ is the similarity threshold.

\subsubsection{Heartbeat Shuffling Augmentation}

Here, we generate positive pairs using \emph{heartbeat shuffling}, an augmentation which distorts the low-level ECG signal while preserving its semantics. Let $\{ t_1, t_2, \dots, t_n \}$ denote the R-peak indices in $\mathbf{x}_i$. For each heartbeat, define the corresponding time index set as:
\[
\mathcal{T}_j = \{ t \in \mathbb{Z} \mid t_j \leq t < t_{j+1} \}, \quad j = 1, \dots, n - 1.
\]
The $j$-th heartbeat segment is then defined as the submatrix:
\[
\mathbf{b}_j = \mathbf{x}_i^{(j)} = \left[ x_{c,t} \right]_{c=1,\dots,C}^{t \in \mathcal{T}_j} \in \mathbb{R}^{C \times |\mathcal{T}_j|}.
\]

We randomly permute the set $\{ \mathbf{x}_i^{(1)}, \dots, \mathbf{x}_i^{(n-1)} \}$ using a permutation $\pi$ over $\{1, \dots, n-1\}$, and construct the augmented segment by concatenating the permuted beats:
\[
\mathbf{x}_i^{\text{shuffle}} = \mathbf{x}_i^{(\pi(1))} \; \| \; \mathbf{x}_i^{(\pi(2))} \; \| \; \dots \; \| \; \mathbf{x}_i^{(\pi(n-1))},
\]
where $\|$ denotes concatenation along the temporal axis.

This augmentation preserves intra-beat morphology while disrupting inter-beat temporal structure. The resulting shuffled view defines an additional positive sample:
\[
\mathcal{P}_i^{\text{aug}} = \{ \mathbf{x}_i^{\text{shuffle}} \}.
\]

\begin{figure}[!t]
    \centerline{\includegraphics[width=\columnwidth]{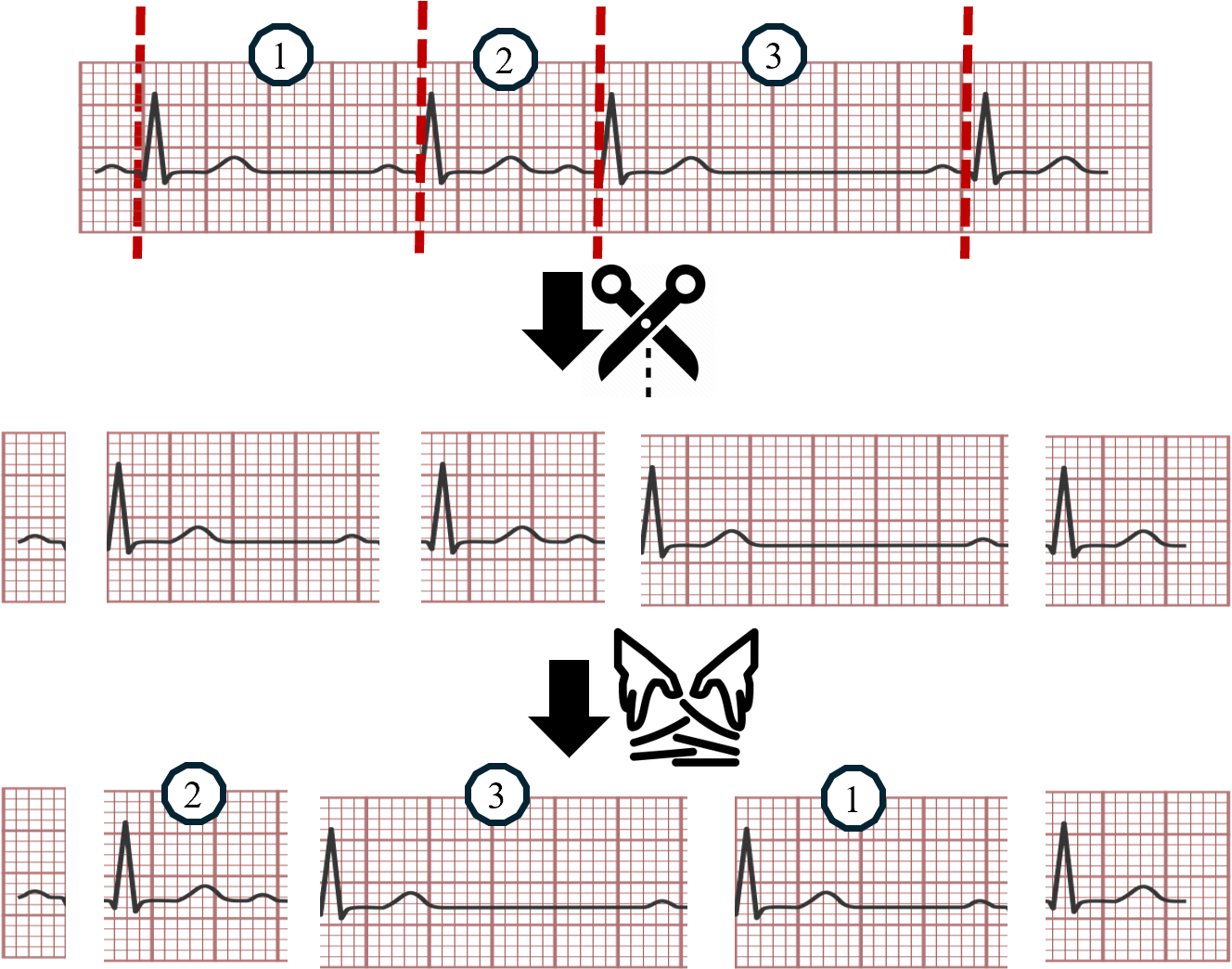}}
    \caption{
Heartbeat Shuffling Augmentation. In this augmentation technique, R-peak onsets are identified to segment the ECG into individual heartbeats. The complete heartbeats are then shuffled in order, maintaining the temporal structure within each heartbeat while altering their overall sequence.
    }
    \label{fig:heartbeat_shuffling}
\end{figure}

Fig.~\ref{fig:heartbeat_shuffling} illustrates the heartbeat-shuffling augmentation process. The first row shows a sample ECG segment, where periods between consecutive R-peaks are identified. In the second row, these periods are segmented into individual heartbeats, and in the third row, the complete heartbeat segments are randomly shuffled.

\subsubsection{Contrastive Loss Computation}
The complete set of positives for each anchor $\mathbf{x}_i$ is formed by aggregating the three previously defined mechanisms:
\[
\mathcal{P}_i = \mathcal{P}_i^{\text{pt}} \cup \mathcal{P}_i^{\text{feat}} \cup \mathcal{P}_i^{\text{aug}}.
\]

Negative samples are defined as all other elements in the batch that are not selected as positives. Since the feature-level selection provides an explicit disjoint separation of the batch based on the cosine similarity threshold $\delta$, we use:
\begin{equation} \label{eqn:physio_feat}
\mathcal{N}_i = \mathcal{N}_i^{\text{feat}} = \left\{ \mathbf{x}_k \mid \text{sim}(\tilde{\mathbf{z}}_i, \tilde{\mathbf{z}}_k) < \delta \right\}.
\end{equation}

For any example $\mathbf{x}_{k}$, let $\mathbf{h}_{k}$ denote its corresponding embedding, i.e., $\mathbf{h}_k = f_\theta(\mathbf{x}_k)$. The contrastive loss for anchor $\textbf{x}_i$ is defined as:
\begin{multline}
\mathcal{L}_i^{\text{contrastive}} = - \frac{1}{|\mathcal{P}_i|} 
\sum_{\mathbf{x}_j \in \mathcal{P}_i} 
\log \\ \Bigg(
\frac{
\exp\left( \text{sim}(\mathbf{h}_i, \mathbf{h}_j)/\tau \right)
}{
\exp\left( \text{sim}(\mathbf{h}_i, \mathbf{h}_j)/\tau \right) 
+ \sum_{\mathbf{x}_k \in \mathcal{N}_i} 
\exp\left( \text{sim}(\mathbf{h}_i, \mathbf{h}_k)/\tau \right)
}
\Bigg)
\end{multline}

The total contrastive loss is the average across all anchors in the batch:
\[
\mathcal{L}_{\text{contrastive}} = \frac{1}{B} \sum_{i=1}^B \mathcal{L}_i^{\text{contrastive}}.
\]

\subsection{Reconstruction Loss}
To encourage preservation of signal structure, we introduce a decoder $g_\phi$ and compute reconstruction loss on both global and peak-centered views. Let $\hat{\mathbf{x}}_i = g_\phi(f_\theta(\mathbf{x}_i))$ be the reconstruction.

\paragraph{Global Loss}
We minimize mean squared error between input and reconstruction:
\[
\mathcal{L}_{\text{global}} = \frac{1}{B} \sum_{i=1}^B \left\| \mathbf{x}_i - \hat{\mathbf{x}}_i \right\|_2^2.
\]

\paragraph{Peak-Based Loss}

To identify peaks in both the input and reconstructed signals, we first apply a 100~ms moving average filter for smoothing and noise reduction, preserving clinically relevant features. Next, local maxima are detected using a prominence threshold. Detected peaks form $\mathbf{x}_i^{\text{peaks}}$ and $\hat{\mathbf{x}}_i^{\text{peaks}}$, which are zero-padded to ensure consistent length. The total peak-based loss is then computed as:
% We extract fixed-width windows centered at detected R-peaks to create $\mathbf{x}_i^{\text{peaks}}$ and $\hat{\mathbf{x}}_i^{\text{peaks}}$, and compute:
\[
\mathcal{L}_{\text{peaks}} = \frac{1}{B} \sum_{i=1}^B \left\| \mathbf{x}_i^{\text{peaks}} - \hat{\mathbf{x}}_i^{\text{peaks}} \right\|_2^2.
\]

\paragraph{Total Reconstruction Loss}
The final reconstruction objective is:
\[
\mathcal{L}_{\text{recon}} = \alpha \mathcal{L}_{\text{global}} + \beta \mathcal{L}_{\text{peaks}},
\]
where $\alpha$ and $\beta$ are weighting coefficients.
\begin{figure}[!t]
    \centering
    \resizebox{0.7\columnwidth}{!}{\includegraphics{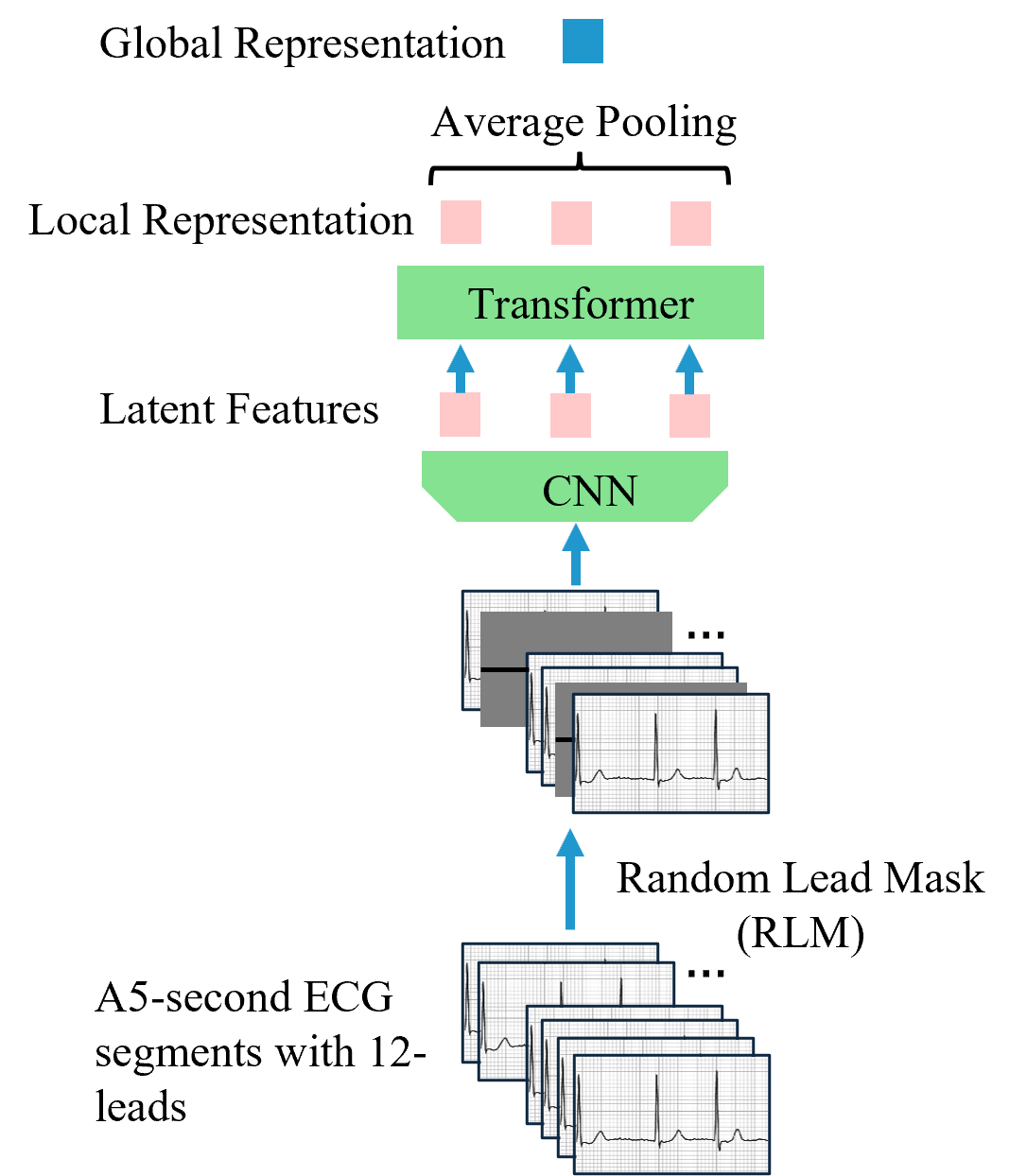}}
    \caption{
The Encoder Architecture is inspired by the model proposed by Oh et al. \cite{Oh2022}. Each lead from the 12-lead, 5-second input ECG segment can be randomly masked. Latent features are then extracted using a CNN layer. Subsequently, transformers generate local representations, and global representations are produced by averaging the local representations. 
    }
    \label{fig:architecture}
\end{figure}
\subsection{Network Architecture}

Our encoder architecture follows the model introduced by Oh et al.~\cite{Oh2022}, which combines a four-block convolutional frontend (each block with 256 channels, stride 2, and kernel size 2) with a 12-layer transformer backbone (hidden size 768, 12 attention heads, feed-forward dimension 3,072). In addition, we adopt their Random Lead Masking (RLM) strategy. In RLM, a random subset of ECG leads is masked during training, encouraging the model to learn lead-invariant representations. This enables transfer from 12-lead datasets used in pretraining to settings with fewer leads. To support downstream tasks, we extend this encoder with a lightweight three-layer decoder (768$\rightarrow$256$\rightarrow$output) for signal reconstruction. A separate linear classifier head is attached for either binary or multilabel classification, depending on the task. The overall architecture is illustrated in Fig.~\ref{fig:architecture}, which shows how convolutional and transformer layers combine to generate the representations.

% while introducing two key modifications: (1) we remove the quantization and local contrastive components, and (2) we replace the original contrastive objective with our feature-informed pair selection mechanism. 

% Explain the decoder architecture

% \subsection{Finetuning for Downstream Tasks}
% Explain how we attach a classification head to the network and train it using multi-class binary classification using CE for downstream tasks.

\section{Datasets and Experiments Details}

\subsection{Datasets}

\begin{table}[t]
\caption{Summary of Datasets Used in This Study}
\label{tab:dataset_summary}
\setlength{\tabcolsep}{6pt}
\renewcommand{\arraystretch}{1.2}
\centering
{\scriptsize
\begin{tabular}{|p{65pt}|p{25pt}|p{20pt}|p{40pt}|p{25pt}|}
\hline
\textbf{Dataset} & \textbf{Records} & \textbf{Leads} & \textbf{Segment Duration (s)} & \textbf{Country} \\ \hline
\multicolumn{5}{|c|}{\textbf{SSL Pre-Training}} \\ \hline
MIMIC-IV-ECG & 787,677 & 12 & 10 & USA \\ \hline
PTB-XL & 21,837 & 12 & 10 & Germany \\ \hline
\multicolumn{5}{|c|}{\textbf{finetuning (26-class multilabel classification)}} \\ \hline
PTB-XL & 21,837 & 12 & 10 & Germany \\ \hline
Ningbo & 34,905 & 12 & 10 & China \\ \hline
\multicolumn{5}{|c|}{\textbf{Validation (26-class multilabel classification)}} \\ \hline
CPSC \& CPSC-Extra & 10,330 & 12 & 6–60 & China \\ \hline
\multicolumn{5}{|c|}{\textbf{Testing (26-class multilabel classification)}} \\ \hline
Chapman & 10,247 & 12 & 10 & China \\ \hline
Georgia & 10,344 & 12 & 5–10 & USA \\ \hline
\multicolumn{5}{|c|}{\textbf{Testing (Binary classification)}} \\ \hline
KGH Private Dataset & 613 & 4 & 10 & Canada \\ \hline
\end{tabular}
}
\end{table}

To train and evaluate our method, we selected several ECG datasets representing a large number of patients and spanning a range of geographical locations and clinical settings. These datasets are summarized in Table~\ref{tab:dataset_summary}. Specifically, we use the following three datasets: 
\begin{itemize}
    \item \textit{MIMIC-IV-ECG:} A large publicly available unlabeled dataset comprising approximately 800,000 12-lead ECG recordings, each 10 seconds in duration, collected from diagnostic ECGs at Beth Israel Deaconess Medical Center (BIDMC)~\cite{mimiciv,physio3}.
    \item \textit{PhysioNet 2021:} A large publicly available  dataset~\cite{physio1,physio2,physio3} comprising of eight independent ECG databases referred to as CPSC, CPSC-Extra, PTB-XL, Georgia, Ningbo, Chapman, PTB, and St. Petersburg INCART, respectively. Following prior work~\cite{Oh2022, ecgfm}, we use only the first six databases in this study, with PTB and INCART being excluded due to their longer recording durations and smaller sample sizes.
    \item \textit{KGH ICU:} A private dataset was collected from bedside monitors in the 33-bed mixed-use ICU at Kingston Health Sciences Centre (KHSC)~\cite{cnn3}. This dataset is valuable for evaluating model generalization because it features a different number of leads and comes from an ICU environment, which typically involves higher noise and variability due to prolonged monitoring and patient movement. Designed for binary classification of atrial fibrillation, it includes 984 patients with a median recording duration of 11.9 hours. Of these, 613 10-second ECG segments were annotated, with 100 segments labeled as AFib. 
\end{itemize}

%\subsubsection{Pretraining datasets}
\subsubsection{Split Selection}
The dataset was divided into pretraining, finetuning, validation, and testing subsets based on three key criteria: (i) Data suitability, such as excluding unlabeled datasets like MIMIC-IV-ECG from finetuning and testing phases, as label supervision is essential for these tasks; (ii) Consistency with prior work, where exclusion decisions (e.g., removing PTB and INCART from downstream tasks) follow precedent—these datasets contain long ECG recordings with global labels that are not suitable for short 10-second segments, since applying such coarse labels introduces substantial label noise; and (iii) robust generalization, ensured by constructing an independent test set that spans multiple geographic regions and clinical environments.

Accordingly, we used the large unlabeled MIMIC-IV-ECG dataset and PTB-XL for pretraining. For finetuning, we used PTB-XL and Ningbo, representing European and Asian populations. CSPC and CSPC-Extra were used for validation and hyperparameter tuning. For testing, we used Chapman, Georgia, and our private KGH dataset (collected in ICU settings in Canada) to evaluate generalizability across diverse populations and clinical contexts.

\begin{table*}[h]
\caption{Performance Comparison of our proposed method and the baselines for different Test datasets}
\label{tab:methods_comparison}
\setlength{\tabcolsep}{1pt}
\setlength{\tabcolsep}{3pt}
\renewcommand{\arraystretch}{1.2}
\centering
{\scriptsize
\begin{tabular}{|p{30pt}|p{60pt}|cc|cc|cccc|}
\hline
\multirow{2}{*}{\textbf{}} & \multirow{2}{*}{\textbf{Method}} & \multicolumn{2}{c|}{\textbf{Chapman}} & \multicolumn{2}{c|}{\textbf{Georgia}} & \multicolumn{4}{c|}{\textbf{KGH}}\\ \cline{3-10} 
 &  &  {AUROC} & {Challenge-Metric}& {AUROC}& {Challenge-Metric}& {Precision}& {Recall}& {F1-Score}& {AUROC} \\ \hline
& Supervised&  0.803&  0.617&  0.724& 0.568&  0.741&  0.845&  0.789& 0.883\\ \hline
\multirow{4}{*}{SSL} & SimCLR~\cite{chen2020}&  0.704&  0.585&  0.681&  0.539&  0.641&  0.715&  0.675& 0.732\\   
 & CLOCS~\cite{Kiyasseh2021}&  0.806&  0.646&  0.711&  0.559&  0.664&  0.721&  0.691& 0.768\\  
 &  {\tiny W2V+CMSC+RLM} \cite{Oh2022}&  0.821&  0.651&  0.729&  0.561&  0.752&  0.892&  0.816& 0.901\\ 
 & ECG-FM \cite{ecgfm}&  - &  - &  - & - &  0.368 &  \textbf{0.971} & 0.523 & 0.861\\ 
 % & ECG-FM \cite{ecgfm}& 0.832& 0.659& 0.761& \textbf{0.595}&  0.760&  0.890&  0.820& 0.917\\   
 & \textbf{PhysioCLR (Ours)} &  \textbf{0.856}&  \textbf{0.663}&  \textbf{0.776}&  \textbf{0.593}&  \textbf{0.771}&  {0.902}&  \textbf{0.831}& \textbf{0.922}\\ \hline
\end{tabular}
}
\end{table*}

\subsubsection{Preprocessing}

\paragraph{Segment Length Filtering}  
All ECG recordings shorter than 10 seconds are discarded. Longer recordings are split into non-overlapping 10-second segments for consistency with pretraining and normalized by z-score normalization. 

\paragraph{ECG-Derived Features}  
We extract physiologically informed features using the \texttt{NeuroKit2} and \texttt{pyHRV} libraries. These features include waveform peak counts, peak amplitudes, peak intervals, heart rate variability metrics, slopes, and energy. The resulting feature vectors are zero-padded to 150 dimensions and reduced to 50 dimensions via PCA. These are used for similarity-based pair selection during contrastive training.

% \textit{Preprocessing} For dataset preprocessing and baseline replication, we used the Fairseq-Signals framework~\cite{fairseq_signals}, a library of deep learning models for ECG data processing built on Facebook's Fairseq framework~\cite{ott2019fairseq}. We applied the same preprocessing steps as described in~\cite{Oh2022, ecgfm}. For SimCLR, we adopted augmentations from \cite{mehari2022}, which include noise addition, baseline wandering, and signal shifts.

\subsection{Experiments}
We designed several experiments testing the efficacy of our methodology to learn a strong ECG encoder that can effectively transfer to downstream clinical tasks. Specifically, we designed the following set of experiments.

\subsubsection{State-of-the-art Comparison}

We compare our method against a diverse set of baselines that represent the current state of self-supervised and supervised ECG learning: 
\begin{itemize}
    \item \textit{Supervised Baseline:} A model trained from scratch using the labeled PhysioNet 2021 corpus, serving as a reference point to quantify the impact of pretraining.
    \item \textit{Contrastive Learning Methods: } SimCLR~\cite{chen2020}, a general-purpose contrastive learning framework, and CLOCS~\cite{Kiyasseh2021}, which introduces a temporal contrastive objective tailored to patient-level ECGs.
    \item \textit{Wav2Vec-based architecture: } The model introduced by Oh et al.~\cite{Oh2022}, incorporating a convolutional transformer encoder, Random Lead Masking (RLM), and the CMSC alignment loss. We refer to this configuration as W2V+CMSC+RLM.
    \item \textit{Foundation Model:} ECG-FM~\cite{ecgfm}, a recently published ECG foundation model that follows the same methodology as W2V+CMSC+RLM but featuring a much larger private dataset. However, since the latest version of ECG-FM was pretrained using our public test datasets, we only report its performance on our private dataset to ensure a fair comparison.
    % \item \textit{Foundation Model:} ECG-FM~\cite{ecgfm}, a large-scale ECG foundation model using the same methodology as W2V+CMSC+RLM but featuring a much larger private training dataset. 
\end{itemize}

\subsubsection{Ablation Studies}
For a more fine-grained assessment of our methodology, we designed several ablation studies. These include (i) the evaluation of our method under reduced amounts of labeled data to further stress-test its robustness to the label-scarce training regime; (ii) a detailed analysis of the sensitivity of our physiological feature-based sampling approach to the choice of threshold; and (iii) an ablation of individual components of our methodology, our novel physiological feature-level sampling (abbreviated as PhysioFeat), heartbeat shuffling augmentation (abbreviated as HRShuff), and reconstruction loss (abbreviated as ReconLoss), to determine their importance. 

%To further understand the contributions of each component in our framework, we designed a set of controlled ablation studies. These include: (1) reducing the amount of labeled data to assess robustness under limited supervision; (2) varying the cosine similarity threshold used in feature-level pair selection to examine sensitivity to contrastive sampling; and (3) evaluating the effect of proposed augmentations such as heartbeat shuffling and the use of reconstruction loss.

\subsection{Evaluation Protocol}
We evaluate the quality of our self-supervised pretraining methodology by assessing the performance of the finetuned model on downstream tasks. Following pretraining, we finetune the model using supervised learning on the finetuning datasets with a 26-class multilabel classification task. Finally, we evaluate this model on the test sets. For the Chapman and Georgia datasets, which are part of PhysioNet, we report multilabel AUROC as well as the CinC 2021 Challenge metric~\cite{physio2}, which penalizes clinically significant misclassifications more heavily. For the KGH dataset, which involves a binary classification task, we report binary AUROC, precision, recall, and F1-score.

Note that evaluating the model on the KGH dataset requires minor post hoc modifications to adapt the 26-class multilabel classifier to a binary Normal vs. AFib classification task. Specifically, we remap the 26 output labels of the network to binary labels as follows: ‘SR’, ‘SA’, ‘SB’, and ‘STach’ are mapped to normal rhythm, while ‘AF’ and ‘AFib’ are mapped to atrial fibrillation.

%\subsubsection{Multilabel Evaluation on PhysioNet}  
%We evaluate model performance on the Chapman and Georgia test sets using AUROC and the CinC 2021 Challenge-metric~\cite{physio2}, which penalizes clinically significant misclassifications more heavily.

%\subsubsection{Binary Classification on KGH Dataset}  
%To evaluate generalization to a different clinical setting, we assess the model on the KGH dataset from a Canadian ICU, where the task is to classify AFib versus normal rhythm. We report AUROC, precision, recall, and F1-score for this binary task. To enable binary evaluation, PhysioNet labels were mapped as follows: ‘SR’, ‘SA’, ‘SB’, and ‘STach’ to normal rhythm; ‘AF’ and ‘AFib’ to atrial fibrillation.

\subsection{Implementation Details}

%\subsubsection{Pretraining}  
\textit{Pretraining:} Pretraining is conducted for 200 epochs using the Adam optimizer with a learning rate of 5e-5 and a batch size of 128. The objective includes both local and global contrastive losses applied to transformer representations. We set the pair selection threshold to 0.25, with reconstruction loss weights $\alpha=0.2$ and $\beta=0.1$. \textit{Finetuning:} For downstream tasks, we finetune the model using a binary cross-entropy loss over 64 epochs with a reduced learning rate of 1e-6. Finetuning is performed on the labeled subset of PhysioNet 2021 and the KGH dataset. \textit{Hyperparameter tuning:} All hyperparameters were tuned to maximize the validation AUROC. For most model and training hyperparameters (e.g., learning rate, optimizer, architecture), we adopted default settings from prior work~\cite{Oh2022,ecgfm}. For parameters specific to our method including physiological similarity threshold ($\delta$ in (~\ref{eqn:physio_feat})), loss weighting terms, and PCA reducted dimensionality, we used the following procedure: first, the parameters were tuned by pretraining and finetuning on a smaller subset of the full pretraining and finetuning set. The pair selection threshold and reconstruction loss weight were identified as the most sensitive hyperparameters; these were subsequently finetuned using the full datasets. \textit{Hardware and Software:} The full code and experimental configurations will be made available at~\url{https://github.com/nooshinmaghsoodi/PhysioCLR} upon acceptance. The PyTorch framework was used together with the Fairseq-Signals library ~\cite{fairseq_signals}. Pretraining took approximately 12 days on 4 NVIDIA A40 GPUs (48 GB each), while finetuning took about 2 days on the same GPU setup. 

%\subsubsection{finetuning}  

\section{Results and Discussion}
 Our results highlight how PhysioCLR advances ECG-based arrhythmia detection by learning robust representations that can generalize across datasets.

\subsection{Baseline Comparison}

The quantitative performance of PhysioCLR and baseline methods on each dataset is detailed in Table~\ref{tab:methods_comparison}. We evaluate PhysioCLR in two distinct clinical scenarios: (i) multilabel classification on PhysioNet 2021 (Chapman and Georgia datasets), and (ii) AFib classification of ECGs from the KGH ICU dataset.

\subsubsection{PhysioCLR Outperforms State-of-the-art Methods on PhysioNet 2021}

On the Chapman dataset, PhysioCLR achieves the highest AUROC of 0.856 and Challenge metric of 0.663, outperforming the best baseline, W2V+CMSC+RLM, which obtains an AUROC of 0.821 and a Challenge metric of 0.651. On the Georgia dataset, PhysioCLR also attains the top AUROC of 0.776 and Challenge-metric of 0.593, ahead of W2V+CMSC+RLM (0.729 and 0.561). 

These results demonstrate that incorporating clinically informed contrastive objectives, including physiological similarity-based pair selection and ECG-specific augmentations, allows PhysioCLR to learn robust and discriminative representations from unlabeled data. The method consistently outperforms supervised training, improving AUROC from 0.803 to 0.856 and Challenge-metric from 0.617 to 0.663 on Chapman, even without access to large labeled datasets. This ability to generalize across patient populations and diagnostic classes highlights the strength of leveraging physiologically meaningful self-supervised learning, especially valuable in real-world clinical settings where labeled data is scarce or heterogeneous.

\subsubsection{PhysioCLR Generalizes Robustly to Noisy ICU ECGs}

On the KGH dataset, which consists of 4-lead ECGs from an ICU setting, PhysioCLR achieves the best performance across several metrics: AUROC of 0.922, F1-score of 0.831, recall of 0.902, and precision of 0.771. Compared to self-supervised baselines such as SimCLR (AUROC 0.732, F1-score 0.675) and CLOCS, PhysioCLR shows substantial improvements. Even against the stronger ECG-specific method, W2V+CMSC+RLM, (AUROC 0.901, F1-score 0.816), it performs better. ECG-FM, a foundation model pretrained solely on ECG data, achieved higher recall; however, our method significantly outperformed it on other metrics, including AUROC (0.861) and F1-score (0.523).

These performance gains are particularly notable given the clinical challenges posed by KGH, including low-lead and noisy input signals. PhysioCLR’s ability to outperform methods suggests that it is well-suited for noisy and resource-constrained settings. This robustness makes PhysioCLR a promising candidate for deployment in environments such as bedside monitoring, where ECG quality is often limited.

\subsection{Ablation Studies}
\label{sec:ablation}
 
\subsubsection{PhysioCLR Mitigates Performance Drop Under Label Scarcity}
Table~\ref{tab:dataset_size_ablation} illustrates the results comparing our method and the supervised method when the number of datasets is decreased gradually from three datasets at the top to just one at the bottom. As shown in Table~\ref{tab:dataset_size_ablation}, PhysioCLR consistently outperforms supervised training across all test sets, particularly as labeled data becomes scarce. With access to all labeled datasets (PTB-XL, Ningbo, CPSC), PhysioCLR achieves AUROC scores of 0.856 (Chapman), 0.776 (Georgia), and 0.922 (KGH), compared to 0.803, 0.724, and 0.883 for the supervised model.

Even when training with only CPSC, PhysioCLR maintains strong performance (0.839 Chapman, 0.732 Georgia, 0.889 KGH), while the supervised baseline suffers greater degradation. The largest gap emerges on the Georgia dataset when PTB-XL is excluded. PhysioCLR's AUROC drops from 0.776 to 0.741, while the supervised model drops from 0.724 to 0.667.

These results highlight the value of self-supervised pretraining for improving robustness to label scarcity. While performance declines as labeled data is removed, the drop is modest and less severe than for supervised models. This illustrates the benefits of contrastive pretraining in learning transferable ECG representations that generalize across domains and demographics.

\begin{table}[h]
\caption{
Comparison of AUROC between PhysioCLR and supervised training across different sizes of labeled finetuning datasets. As the amount of labeled data decreases, PhysioCLR maintains strong performance across all test sets—Chapman, Georgia, and KGH—while the supervised model’s performance drops more significantly.
% AUROC comparison between PhysioCLR and supervised training as the number of labeled finetuning datasets decreases.
}

\label{tab:dataset_size_ablation}
\setlength{\tabcolsep}{1pt}
\setlength{\tabcolsep}{3pt}
\renewcommand{\arraystretch}{1.2}
\centering
{\scriptsize
\begin{tabular}{|p{75pt}|ccc|ccc|}
\hline
 \multirow{2}{*}{\textbf{Labeled Datasets}}& \multicolumn{3}{c|}{\textbf{PhysioCLR}}  & \multicolumn{3}{c|}{\textbf{Supervised}}\\  \cline{2-7} 
 &  Chapman& Georgia& KGH& Chapman& Georgia& KGH \\ \hline
 CPSC&  0.839&  0.732&  0.889&  0.716&  0.632& 0.821\\
 Ningbo+CPSC&  0.851&  0.741&  0.903&  0.772&  0.667& 0.851\\   
 PTB-XL+Ningbo+CPSC&  0.856&  0.776&  0.922&  0.803&  0.724& 0.883\\  \hline
\end{tabular}
}
\end{table}

% \subsubsection{Similarity Threshold Encodes Domain Knowledge Critical for Effective Pair Selection}
\subsubsection{Physiological Similarity Improves Positive Pair Selection}

\begin{figure}[!t]
    \centerline{\includegraphics[width=\columnwidth]{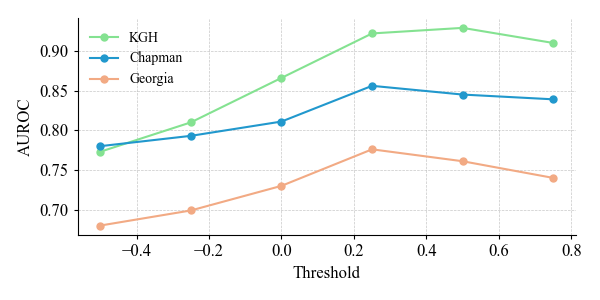}}
    \caption{
Impact of similarity threshold on AUROC performance metric for PhysioNet 2021 and KGH datasets. The cosine similarity thresholds determine the number of positive and negative pairs. Lower thresholds increase the number of positive pairs, while higher thresholds increase the proportion of hard negatives.
    }
    \label{fig:threshold_selection}
\end{figure}

We evaluated the effect of the cosine similarity threshold in feature-level pair selection by testing values from $-0.5$ to $0.75$. Fig.~\ref{fig:threshold_selection} shows the effect of this threshold on AUROC. As shown in this figure, model performance varies notably across this range, underscoring the importance of how physiological similarity is defined and implemented in self-supervised learning.

On all datasets, performance is low at low thresholds, peaks with a threshold in the range $0.25$ to $0.5$, then gradually declines. On Chapman and Georgia, performance peaks around a threshold of $0.25$ (AUROC 0.84 and 0.76, respectively) and declines gradually at higher thresholds. This suggests that an overly strict positive pair definition limits the model’s ability to capture intra-class variability. Conversely, KGH performance improves up to $0.5$ (AUROC 0.92), consistent with the notion that harder negatives are more beneficial in noisy ICU settings. At very low thresholds such as $-0.5$, performance drops substantially, most likely due to the generation of semantically unrelated false positive pairs.

This analysis highlights the critical role of positive and negative pair definitions in ECG contrastive learning. It supports our central claim that incorporating domain knowledge, in this case, physiological similarity, is essential for effective representation learning. Our findings are consistent with prior work, such as SimCLR and InfoMin~\cite{tian2020makes}, which emphasize the importance of positives being neither ``too similar" nor ``too different". In practice, we observe that a threshold range of 0.25 to 0.5 performs reliably across datasets and can be selected through simple cross-validation.

\subsubsection{All Method Components Contribute to Robust ECG Representation Learning}

\begin{figure}[!t]
    \centerline{\includegraphics[width=\columnwidth]{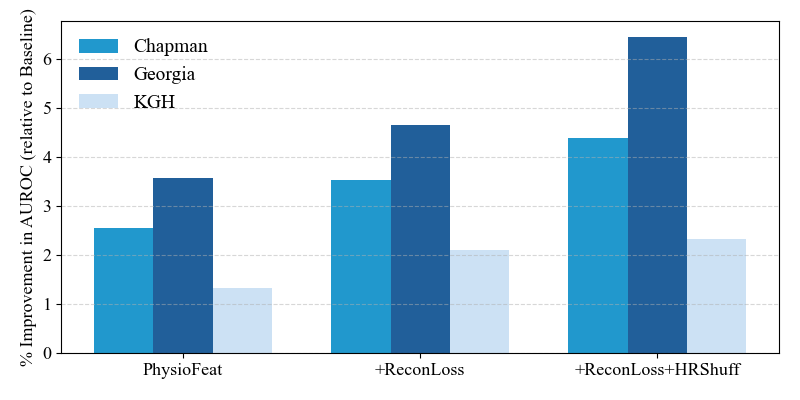}}
    \caption{
Ablation Study on Proposed Method Components: Each group of bars represents the improvement in AUROC over the baseline (\textbf{W2V+CMSC+RLM}) for three evaluation datasets (Chapman, Georgia, KGH). The methods tested include \textbf{PhysioFeat} (Physiological Feature-Level Pair Selection), \textbf{HRShuff} (Heartbeat Shuffling), \textbf{ReconLoss} (the combination of reconstruction and contrastive loss). }
    \label{fig:method_combination}
\end{figure}

Fig.~\ref{fig:method_combination} illustrates the individual and combined contributions of each proposed component. Each group of bars shows the AUROC improvement over the W2V+CMSC+RLM baseline across the test datasets. First, the introduction of PhysioFeat alone yields a substantial improvement of $2.49\%$ AUROC (averaged) across datasets compared to the strong W2V+CMSC+RLM baseline. The addition of ReconLoss yields an additional 9.5\% improvement. With all components combined, the model achieves an average improvement $4.39\%$: 4.0\% for Chapman, 6.5\% for Georgia, and 2.6\% for KGH. These results demonstrate that while each single component is effective in improving representation quality by adding physiological priors from different angles, their incorporation into the unified PhysioCLR framework provides the strongest gains. %Effective ECG representation learning in self-supervised settings requires a unified approach that leverages multiple physiological priors, including carefully selected augmentations, domain-informed pair construction, and hybrid loss objectives.

\section{Conclusion}
 Our work demonstrates the importance of embedding physiological knowledge into self-supervised learning, ultimately supporting more generalizable clinical ECG interpretation for arrhythmia classification. Building upon this motivation, SSL provides a promising future for the analysis of biomedical signals: training deep networks from vast quantities of unlabeled data provides a scalable solution compared to conventional label-dependent methods. Still, domain knowledge of the data and its underlying physiology remains key to unlocking the full potential of these algorithms. We introduce PhysioCLR, a unified framework that incorporates physiological priors into SSL for ECG. PhysioCLR combines contrastive and reconstruction objectives that explicitly reflect the morphological and temporal characteristics of ECG signals, promoting representations aligned with their physiological semantics. Empirical results demonstrate that PhysioCLR learns more robust and transferable features than prior methods, enabling improved performance across multiple downstream clinical tasks. Given its success, there remain opportunities for further refinement and extension. Since the extracted physiological features play a central role in guiding pair selection, enhancing the precision of feature computation is critical for improving representation quality, especially in the presence of noise or complex patterns. Additionally, fixed thresholds for similarity-based pairing may not generalize optimally across all datasets; meanwhile, an extension of this approach to other physiological signals and further refinement of the positive pair selection strategy are also promising next steps.

\section*{Clarifications}
This study was approved by Queen’s University Health Sciences Research Ethics Board—Reference number 6024689. Consent in this study was waived because the data were collected as part of routine care and were stored in a de-identified format.

Stephanie Sibley declares the following conflicts of interest: receipt of honoraria from Think Research; meeting sponsorships from Boston Scientific, Trimedic, and Icentia; and service as a hospital organ-donation physician with the Trillium Gift of Life Network at Ontario Health. No other author has any conflict of interest to disclose.


\begin{thebibliography}{00}

\bibitem{faust2018review}
O.~Faust, Y.~Hagiwara, T.~J.~Hong, R.~S.~Tan, and U.~R.~Acharya,
“Deep learning for healthcare applications based on physiological signals: A review,”
\textit{Comput. Methods Programs Biomed.}, vol.~161, pp.~1--13, 2018.

\bibitem{dinov2}
M. Oquab, T. Darcet, T. Moutakanni, H. Vo, M. Szafraniec, V. Khalidov \textit{et al.},
“DINOv2: Learning robust visual features without supervision,”
\textit{arXiv:2304.07193}, Apr. 2023.

\bibitem{chen2020}
T. Chen, S. Kornblith, M. Norouzi and G. Hinton,
“A simple framework for contrastive learning of visual representations,”
in \textit{Proc. 37th Int. Conf. Machine Learning (ICML)}, Nov. 2020,  
pp. 1597–1607.

\bibitem{byol}
J.-B. Grill, F. Strub, F. Altché \textit{et al.},
“Bootstrap your own latent: A new approach to self-supervised learning,”
\textit{Adv. Neural Inf. Process. Syst.}, vol. 33, pp. 21271–21284, 2020.

\bibitem{reconst1}
X. Chen, M. Ding, X. Wang \textit{et al.},
“Context autoencoder for self-supervised representation learning,”
\textit{Int. J. Comput. Vis.}, vol. 132, no. 1, pp. 208–223, Jan. 2024,  
doi: 10.1007/s11263-023-01852-4.

\bibitem{reconst2}
A. v. d. Oord, Y. Li and O. Vinyals,
“Representation learning with contrastive predictive coding,”
\textit{arXiv:1807.03748}, Jul. 2018.

\bibitem{reconst3}
Y. Pang, W. Wang, F. E. Tay, W. Liu, Y. Tian and L. Yuan,
“Masked autoencoders for point-cloud self-supervised learning,”
in \textit{Proc. ECCV}, Cham, Switzerland: Springer, Oct. 2022,  
pp. 604–621.

\bibitem{mohamed2022review}
A. Mohamed, H.-Y. Lee, L. Borgholt \textit{et al.},
“Self-supervised speech representation learning: A review,”
\textit{IEEE J. Sel. Top. Signal Process.}, vol. 16, no. 6, pp. 1179–1210, Dec. 2022.

\bibitem{augmentationIssue}
A. R. Alkhulaifi et al., “Which augmentation should I use? An empirical investigation of augmentations for self-supervised phonocardiogram representation learning,” in \textit{Proc. Annual International Conference of the IEEE Engineering in Medicine and Biology Society (EMBC)}, pp. 1–4, July 2023.

% I. Nault, N. Lellouche, S. Matsuo \textit{et al.},
% “Clinical value of fibrillatory-wave amplitude on surface ECG in patients with persistent atrial fibrillation,”
% \textit{J. Interv. Card. Electrophysiol.}, vol. 26, no. 1, pp. 11–19, Oct. 2009.

\bibitem{hannun2019cardiologist}
A. Y. Hannun, P. Rajpurkar, M. Haghpanahi \textit{et al.},
“Cardiologist-level arrhythmia detection and classification in ambulatory electrocardiograms using a deep neural network,”
\textit{Nat. Med.}, vol. 25, no. 1, pp. 65–69, Jan. 2019; see also published correction, doi: 10.1038/s41591-019-0359-9.

\bibitem{clifford2017afdb}
G. D. Clifford, C. Liu, B. Moody \textit{et al.},
“AF classification from a short single-lead ECG recording: The PhysioNet/Computing in Cardiology Challenge 2017,”
\textit{Comput. Cardiol.}, vol. 44, pp. 1–4, Sep. 2017.

\bibitem{acharya2017deep}
U. R. Acharya, H. Fujita, S. L. Oh \textit{et al.},
“Deep convolutional neural network for the automated diagnosis of congestive heart failure using ECG signals,”
\textit{Appl. Intell.}, vol. 49, no. 1, pp. 16–27, 2019.

\bibitem{ibtehaz2022ecg}
N. Ibtehaz, M. H. Mahmud and A. B. M. Al Islam,
“ECG segmentation using a deep learning model,”
\textit{Biocybern. Biomed. Eng.}, vol. 42, no. 2, pp. 418–431, 2022.

\bibitem{mehari2022}
T. Mehari and N. Strodthoff,
“Self-supervised representation learning from 12-lead ECG data,”
\textit{Comput. Biol. Med.}, vol. 141, Art. no. 105114, Feb. 2022.

\bibitem{Gopal2021}
B. Gopal, R. Han, G. Raghupathi, A. Ng, G. Tison and P. Rajpurkar,
“3KG: Contrastive learning of 12-lead electrocardiograms using physiologically-inspired augmentations,”
in \textit{Proc. Machine Learning for Health (ML4H)}, PMLR vol. 158, pp. 156–167, Dec. 2021.

\bibitem{Le2023}
D. Le, S. Truong, P. Brijesh, D. A. Adjeroh and N. Le,
“sCL-ST: Supervised contrastive learning with semantic transformations for multiple-lead ECG arrhythmia classification,”
\textit{IEEE J. Biomed. Health Inform.}, vol. 27, no. 6, pp. 2818–2828, Jun. 2023.

\bibitem{chen2021}
H. Chen, G. Wang, G. Zhang, P. Zhang and H. Yang,
“CLECG: A novel contrastive learning framework for electrocardiogram arrhythmia classification,”
\textit{IEEE Signal Process. Lett.}, vol. 28, pp. 1993–1997, Dec. 2021.

\bibitem{Kiyasseh2021}
D. Kiyasseh, T. Zhu and D. A. Clifton,
“CLOCS: Contrastive learning of cardiac signals across space, time and patients,”
in \textit{Proc. 38th ICML}, PMLR vol. 139, pp. 5606–5615, Jul. 2021.

\bibitem{Wang2024}
Y. Wang, Y. Han, H. Wang and X. Zhang,
“Contrast everything: A hierarchical contrastive framework for medical time-series,”
\textit{Adv. Neural Inf. Process. Syst.}, vol. 36, Art. no. 15548, Dec. 2024.

\bibitem{Oh2022}
J. Oh, Y. Lee and J. Kim,
“Lead-agnostic self-supervised learning for local and global representations of electrocardiogram,”
in \textit{Proc. Machine Learning for Health (ML4H)}, PMLR vol. 174, pp. 322–337, Dec. 2022.

\bibitem{chen2024multi}
J. Chen, W. Wu, T. Liu and S. Hong,
“Multi-channel masked autoencoder and comprehensive evaluations for reconstructing 12-lead ECG from arbitrary single-lead ECG,”
\textit{npj Cardiovasc. Health}, vol. 1, no. 1, Art. no. 13, 2024.

\bibitem{na2024guiding}
Y. Na, M. Park, Y. Tae and S. Joo,
“Guiding masked representation learning to capture spatio-temporal relationships of electrocardiogram,”
\textit{arXiv:2402.09450}, Feb. 2024.

\bibitem{Liu2024}
W. Liu, H. Zhang, S. Chang, H. Wang, J. He and Q. Huang,
“Learning representations for multi-lead electrocardiograms from morphology–rhythm contrast,”
\textit{IEEE Trans. Instrum. Meas.}, early access, pp. 1–12, Jan. 2025, doi: 10.1109/TIM.2025.3274458.

\bibitem{Zhou2025}
X. Zhou, M. Shi, X. Yu \textit{et al.},
“Self-supervised inter–intra period-aware ECG representation learning for detecting atrial fibrillation,”
\textit{Biomed. Signal Process. Control}, vol. 100, Art. no. 106939, 2025.

\bibitem{clifford2006advanced}
G. D. Clifford, F. Azuaje and P. E. McSharry,
\textit{Advanced Methods and Tools for ECG Data Analysis}. Norwood, MA, USA: Artech House, 2006.

\bibitem{acharya2017automated}
U. R. Acharya, H. Fujita, S. L. Oh \textit{et al.},
“Automated identification of shockable and non-shockable life-threatening ventricular arrhythmias using convolutional neural network,”
\textit{Future Gener. Comput. Syst.}, vol. 79, pp. 952–959, 2018.

\bibitem{Surawicz}
B. Surawicz and T. Knilans,
\textit{Chou’s Electrocardiography in Clinical Practice}. 6th ed. Amsterdam, The Netherlands: Elsevier, 2008.

\bibitem{svm}
S. Osowski, L. T. Hoai and T. Markiewicz,
“Support vector machine-based expert system for reliable heartbeat recognition,”
\textit{IEEE Trans. Biomed. Eng.}, vol. 51, no. 4, pp. 582–589, Apr. 2004.

\bibitem{svm2}
T. Ince, S. Kiranyaz and M. Gabbouj,
“Automated patient-specific classification of premature ventricular contractions,”
in \textit{Proc. 30th Annu. Int. Conf. IEEE EMBC}, Vancouver, BC, Canada, Aug. 2008, pp. 5474–5477.

\bibitem{svm3}
Q. Zhao and L. Zhang,
“ECG feature extraction and classification using wavelet transform and support vector machines,”
in \textit{Proc. Int. Conf. Neural Networks and Brain}, Beijing, China, Oct. 2005, vol. 2, pp. 1089–1092.

\bibitem{cnn1}
Y. Li, Y. Pang, J. Wang and X. Li,
“Patient-specific ECG classification by deeper CNN from generic to dedicated,”
\textit{Neurocomputing}, vol. 314, pp. 336–346, Nov. 2018.

\bibitem{cnn2}
Z. Chen, D. Yang, T. Cui \textit{et al.},
“A novel imbalanced-dataset mitigation and ECG classification model based on combined 1D CBAM autoencoder and lightweight CNN,”
\textit{Biomed. Signal Process. Control}, vol. 87, Art. no. 105437, 2024.

\bibitem{cnn3}
B. Chen, D. M. Maslove, J. D. Curran \textit{et al.},
“A deep learning model for the classification of atrial fibrillation in critically ill patients,”
\textit{Intensive Care Med. Exp.}, vol. 11, Art. no. 2, Jan. 2023.

\bibitem{rnn}
S. Singh, S. K. Pandey, U. Pawar and R. R. Janghel,
“Classification of ECG arrhythmia using recurrent neural networks,”
\textit{Procedia Comput. Sci.}, vol. 132, pp. 1290–1297, 2018.

\bibitem{lstm}
V. Satheeswaran, G. N. Chandrika, A. Mitra \textit{et al.},
“Deep learning-based classification of ECG signals using RNN and LSTM mechanism,”
\textit{J. Electron. Electromed. Eng. Med. Inform.}, vol. 6, no. 4, pp. 332–342, 2024.

\bibitem{transformer}
Y. Xia, Y. Xu, P. Chen, J. Zhang and Y. Zhang,
“Generative adversarial network with transformer generator for boosting ECG classification,”
\textit{Biomed. Signal Process. Control}, vol. 80, Art. no. 104276, 2023.

\bibitem{transformer2}
H. El-Ghaish and E. Eldele,
“ECGTransForm: Empowering adaptive ECG arrhythmia classification framework with bidirectional transformer,”
\textit{Biomed. Signal Process. Control}, vol. 89, Art. no. 105714, 2024.

\bibitem{ecgfm}
K. McKeen, L. Oliva, S. Masood \textit{et al.},
“ECG-FM: An open electrocardiogram foundation model,”
\textit{arXiv:2408.05178}, May 2025.

\bibitem{dinosr}
A. H. Liu, H.-J. Chang, M. Auli, W.-N. Hsu and J. Glass,
“DinoSR: Self-distillation and online clustering for self-supervised speech representation learning,”
\textit{Adv. Neural Inf. Process. Syst.}, vol. 36, pp. 58346–58362, 2023.

\bibitem{Wang2023}
N. Wang, P. Feng, Z. Ge, Y. Zhou, B. Zhou and Z. Wang,
“Adversarial spatiotemporal contrastive learning for electrocardiogram signals,”
\textit{IEEE Trans. Neural Netw. Learn. Syst.}, vol. 35, no. 10, pp. 13845–13859, Oct. 2024.

\bibitem{huynh2022boosting}
T. Huynh, S. Kornblith, M. R. Walter, M. Maire and M. Khademi,
“Boosting contrastive self-supervised learning with false-negative cancellation,”
in \textit{Proc. IEEE/CVF Winter Conf. Appl. Comput. Vis. (WACV)}, Jan. 2022, pp. 2785–2795.

\bibitem{masked1}
H. Zhang, W. Liu, J. Shi \textit{et al.},
“MaeFE: Masked autoencoders family of electrocardiogram for self-supervised pre-training and transfer learning,”
\textit{IEEE Trans. Instrum. Meas.}, vol. 72, pp. 1–16, 2023.

\bibitem{masked2}
Y. Zhou, X. Diao, Y. Huo \textit{et al.},
“Masked transformer for electrocardiogram classification,”
\textit{arXiv:2309.07136}, Sep. 2023.

\bibitem{rebar}
M. A. Xu, A. Moreno, H. Wei, B. M. Marlin and J. M. Rehg,
“REBAR: Retrieval-based reconstruction for time-series contrastive learning,”
\textit{arXiv:2311.00519}, Nov. 2023.

\bibitem{cmelt}
M. Pham, A. Saeed and D. Ma,
“C-MELT: Contrastive enhanced masked auto-encoders for ECG-language pre-training,”
\textit{arXiv:2410.02131}, Oct. 2024.

\bibitem{neurokit2}
D. Makowski, T. Pham, Z. J. Lau \textit{et al.},
“NeuroKit2: A python toolbox for neurophysiological signal processing,”
\textit{Behav. Res. Methods}, vol. 53, no. 4, pp. 1689–1696, Aug. 2021; erratum, doi: 10.1038/s41597-022-01643-5.

\bibitem{pan1985qrs}
J. Pan and W. J. Tompkins,
“A real-time QRS detection algorithm,”
\textit{IEEE Trans. Biomed. Eng.}, vol. 32, no. 3, pp. 230–236, Mar. 1985.

\bibitem{rlm}
A. Baevski, H. Zhou, A. Mohamed and M. Auli,
“wav2vec 2.0: A framework for self-supervised learning of speech representations,”
\textit{arXiv:2006.11477}, Jun. 2020.

\bibitem{physio3}
A. L. Goldberger, L. A. Amaral, L. Glass \textit{et al.},
“PhysioBank, PhysioToolkit and PhysioNet: Components of a new research resource for complex physiologic signals,”
\textit{Circulation}, vol. 101, no. 23, pp. e215–e220, Jun. 2000.

\bibitem{mimiciv}
B. Gow, T. Pollard, L. A. Nathanson \textit{et al.},
“MIMIC-IV-ECG: Diagnostic electrocardiogram matched subset (version 1.0),”
\textit{PhysioNet}, 2023.

\bibitem{physio1}
M. A. Reyna, N. Sadr, E. A. P. Alday \textit{et al.},
“Will two do? Varying dimensions in electrocardiography: The PhysioNet/Computing in Cardiology Challenge 2021,”
in \textit{Proc. Comput. Cardiol. (CinC)}, Brno, Czech Republic, Sep. 2021, pp. 1–4.

\bibitem{physio2}
E. A. Perez Alday, A. Gu, A. J. Shah \textit{et al.},
“Classification of 12-lead ECGs: The PhysioNet/Computing in Cardiology Challenge 2020,”
\textit{Physiol. Meas.}, vol. 41, no. 12, Art. no. 124003, Jan. 2021.

\bibitem{ott2019fairseq}
M. Ott, S. Edunov, A. Baevski \textit{et al.},
“fairseq: A fast, extensible toolkit for sequence modeling,”
in \textit{Proc. NAACL-HLT 2019 (Demonstrations)}, Minneapolis, MN, USA, Jun. 2019, pp. 48–53.

\bibitem{fairseq_signals}
J. Oh,
“fairseq-signals: Self-supervised learning framework for biosignals (ECG, PPG),” GitHub repository,  
https://github.com/Jwoo5/fairseq-signals, accessed May 21, 2025.

\bibitem{tian2020makes}
Y. Tian, C. Sun, B. Poole, D. Krishnan, C. Schmid and P. Isola,
“What makes for good views for contrastive learning?”
\textit{Adv. Neural Inf. Process. Syst.}, vol. 33, pp. 6827–6839, 2020.

\end{thebibliography}
\end{document}